\documentclass[runningheads]{llncs}
\usepackage{cite}
\usepackage{amsmath,amssymb,amsfonts,mathtools,enumerate}
\usepackage{algorithmic}
\usepackage{graphicx}
\usepackage{textcomp}
\usepackage{xcolor}
\usepackage{hyperref}
\usepackage[ruled,vlined]{algorithm2e}
\usepackage{xcolor}
\usepackage{tikz}
\usetikzlibrary{shapes,arrows,shadows}
\usepackage{url}
\newcommand{\R}{\mathds{R}}

\newcommand{\bR}{\bar{\mathds{R}}}

\newcommand{\vetx}{\boldsymbol{x}}

\newcommand{\bpm}{\begin{bmatrix}}
\newcommand{\epm}{\end{bmatrix}}

\newcommand{\alphav}{\boldsymbol{\alpha}}

\newcommand{\bp}{\mbox{\boldmath$p$}}
\newcommand{\bq}{\mbox{\boldmath$q$}}

\newcommand{\betav}{\boldsymbol{\beta}}

\newcommand{\qeq}{\quad \mbox{and} \quad}

\DeclareMathOperator*{\argmax}{arg\,max}

\def\BibTeX{{\rm B\kern-.05em{\sc i\kern-.025em b}\kern-.08em
    T\kern-.1667em\lower.7ex\hbox{E}\kern-.125emX}}

\usepackage{dsfont}
\usepackage{booktabs}
\usepackage{enumerate}
\usepackage{adjustbox}
\usepackage{amssymb,amsfonts}

\everymath{\displaystyle}
\newcommand{\ba}{{\bf a}}
\newcommand{\bb}{{\bf b}}

\renewcommand{\bp}{{\bf p}}
\renewcommand{\bq}{{\bf q}}
\newcommand{\bm}{{\bf m}}
\newcommand{\bw}{{\bf w}}
\newcommand{\bx}{{\bf x}}

\newcommand{\bxi}{\boldsymbol{\xi}}

\newcommand{\rip}{{\rangle}}
\newcommand{\lip}{{\langle}}

\SetKwInOut{KwInput}{Input}
\SetKwInOut{KwOutput}{Output}
\usepackage[ruled]{algorithm2e}
\begin{document}

\title{Least-Squares Linear Dilation-Erosion Regressor Trained using a Convex–Concave Procedure\thanks{This work was supported in part by CNPq under grant no. 315820/2021-7, FAPESP under grant no. 2022/01831-2, and Coordenação de Aperfeiçoamento de Pessoal de Nível Superior - Brasil (CAPES) - Finance Code 001.}
}
\titlerunning{Least-Squares Linear Dilation-Erosion Regression}
     \author{Angelica Lourenço Oliveira\orcidID{0000-0002-8689-8522} \and \\
     Marcos Eduardo Valle\orcidID{0000-0003-4026-5110}
     }
     \authorrunning{A.L. Oliveira and M.E. Valle}
     \institute{University of Campinas, Campinas -- São Paulo, Brazil. \\
     \email{ra211686@ime.unicamp.br} and \email{valle@ime.unicamp.br}}
\maketitle     
\begin{abstract}
This paper presents a hybrid morphological neural network for regression tasks called linear dilation-erosion regressor ($\ell$-DER). An $\ell$-DER is given by a convex combination of the composition of linear and morphological operators. They yield continuous piecewise linear functions and, thus, are universal approximators. Besides introducing the $\ell$-DER model, we formulate their training as a difference of convex (DC) programming problem. Precisely, an $\ell$-DER is trained by minimizing the least-squares using the convex-concave procedure (CCP). Computational experiments using several regression tasks confirm the efficacy of the proposed regressor, outperforming other hybrid morphological models and state-of-the-art approaches such as the multilayer perceptron network and the radial-basis support vector regressor.

\end{abstract}

\section{Introduction}

Dilations and erosions are the elementary operations of mathematical morphology, a non-linear theory widely used for image processing and analysis \cite{heijmans_mathematical_1995,soille_morphological_1999}. In the middle 1990s, Ritter et al. proposed the first morphological neural networks whose processing units, the morphological neurons, perform dilations and erosions \cite{ritter96c,Ritter1998MorphologicalMemories}. Morphological neurons are obtained by replacing the usual dot product with either the maximum or minimum of sums. Because of the maximum and minimum operations, morphological neural networks are usually cheaper than traditional models. However, training morphological neural networks are often a big challenge because of the non-differentiability of the maximum and minimum operations \cite{pessoa00}. This paper addresses this issue by proposing a different method for training a hybrid morphological neural network for regression tasks. Precisely, we focus on training the so-called linear dilation-erosion perceptron using a difference of convex (DC) optimization method.

A dilation-erosion perceptron (DEP) is a hybrid morphological neural network obtained by a convex combination of dilations and erosions \cite{araujo_class_2011}. Despite its application in regression tasks such as time-series prediction and software development cost estimation \cite{araujo_class_2011,Araujo2012AnEstimation}, the DEP model has an inherent drawback: as an increasing operator, it implicitly assumes an ordering relationship between inputs and outputs \cite{valle_reduced_2020}. Fortunately, one can circumvent this problem by adding neurons that perform anti-dilations or anti-erosions \cite{sussner_morphological_2011}. For example, considering the importance of dendritic structures, Ritter and Urcid presented a single morphological neuron that circumvents the limitations of the DEP model \cite{ritter_lattice_2003}. 

Alternatively, Valle proposed the reduced dilation-erosion perceptron ($r$-DEP) using concepts from multi-valued mathematical morphology \cite{valle_reduced_2020}. In a few words, an $r$-DEP is obtained by composing an appropriate transformation with the DEP model, i.e., the inputs are transformed before they are fed to the DEP model. However, choosing the proper transformation is challenging in designing an efficient $r$-DEP model. As a solution, Oliveira and Valle proposed the so-called linear dilation-erosion perceptron ($\ell$-DEP) by considering linear mappings instead of arbitrary transformations \cite{oliveira_linear_2021}. Interestingly, the linear dilation-erosion perceptron is equivalent to a maxout network with two hidden units \cite{goodfellow_maxout_2013}. The $\ell$-DEP model is also closely related to one of two hybrid morphological neural networks investigated by Hernández et al. for big data classification \cite{Hernandez2020HybridClassification}. 

From a mathematical point of view, the $\ell$-DEP yields a continuous piecewise linear function. Thus, like many traditional neural networks, they are universal approximators; that is, an $\ell$-DEP model can approximate a continuous function within any desired accuracy in a compact region in a Euclidean space \cite{wang_general_2004}. Besides the models mentioned above, the morphological/linear perceptron \cite{sussner_extreme_2020} and the hybrid multilayer morphological network \cite{mondal_morphological_2020} also exhibit universal approximation capability. However, all these hybrid morphological networks differ significantly in the training rule.

Like traditional neural networks, maxout and hybrid multilayer morphological networks can be trained using the stochastic gradient descent (SGD) method \cite{goodfellow_maxout_2013,mondal_morphological_2020}. Despite the non-differentiability of morphological operators, Henández et al. also used the SDG method for training their hybrid morphological neural networks \cite{Hernandez2020HybridClassification}. In contrast, Sussner and Campiotti circumvented the non-differentiability of the morphological operators using an extreme learning machine approach to train the morphological/linear perceptron \cite{sussner_extreme_2020}. Apart from the hybrid morphological/linear network literature, Ho et al. formulated the learning of a continuous piecewise linear function as a difference of convex functions (DC) programming problem \cite{ho_dca-based_2021}. Because the $\ell$-DEP model can be identified with continuous piecewise linear functions, it can also be trained using DC programming. Indeed, this paper uses the concave-convex procedure (CCP) for training the $\ell$-DEP model for regression tasks. Interestingly, the CCP results in more straightforward optimization problems than the approach of Ho et al.. Computational experiments with many regression tasks confirm the advantage of the proposed $\ell$-DEP model against the hybrid morphological/linear methods from the literature and the approach based on the DC optimization of Ho et al. 

The paper is organized as follows. The following section reviews some basic concepts regarding DC optimization, including definitions and the CCP optimization technique. Section \ref{sec:lder} presents the $\ell$-DEP model  for regression tasks while the training based on CCP is addressed in Section \ref{sec:approaches}. Computational experiments using several regression problems from the literature are detailed in Section \ref{sec:experiments}. The paper finishes with some remarks in Section \ref{sec:concluding}.

\section{Difference of Convex Optimization and the Convex-Concave Procedure}\label{sec:basic}

The difference of convex (DC) functions results non-convex functions that enjoy interesting and useful properties \cite{hartman_functions_1959,tuy_dc_2016}. DC optimization aims to optimize such kinds of functions. Applications of DC optimization programs include signal processing, machine learning, computer vision, and statistics \cite{lipp_variations_2015,shen_disciplined_2016}. This section presents some basic concepts of DC optimization. This section also addresses the convex-concave procedure proposed by Yuille and Rangarajan for constrained DC optimization problems \cite{yuille_concave-convex_2003}. The convex-concave procedure will be used in Section \ref{sec:approaches} for training the linear dilation-erosion perceptron for regression tasks.

\subsection{Basic Concepts of DC Optimization}

Consider a real-valued function $ f: \mathcal{C} \rightarrow \mathds{R} $ defined in a convex set $ \mathcal{C} \subseteq \R^n$. We say that $f$ is a DC function if there are convex functions $g, h: \mathcal{C} \rightarrow \mathds{R} $ such that \begin{equation}  \label{eq:DC-decomposition}
f(\alphav) = g(\alphav) - h(\alphav), \quad \forall \alpha \in \mathcal{C}.
\end{equation}
Recall that a function is convex if the line joining two points on its graph is below the graph \cite{luenberger_linear_1984}. The functions $g$ and $h$ are called the DC components of $f$ and the identity in \eqref{eq:DC-decomposition} is is the DC decomposition of $f$. 

In a DC optimization problem, the objective and constraints are  DC functions. In this paper, we focus on the following constrained DC optimization problem: 
\begin{equation}\label{eq:dccp}
\begin{cases}
    \mathop{\text{minimize}}_{\alphav \in \mathds{R}^n} & \; f_0(\alphav) = g_0(\alphav)-h_0(\alphav) \\ 
    \text{subject to}  & \; f_i(\alpha) = g_i(\alphav)-h_i(\alphav) \le 0, \quad i =1,\ldots,m,
\end{cases}
\end{equation}
where $g_i:\mathds{R}^n \to \mathds{R}$ and $h_i:\mathds{R}^n \to \mathds{R}$ are convex functions for all $i=0,1,\ldots,m$. 

In general terms, DC optimization methods take advantage of the convexity of the DC components $g_i$ and $h_i$ of $f_i$. For example, many methods approximate the convex terms $h_i$ of the DC decomposition of $f_i$ an affine function, resulting in convex optimization subproblems that are solved more effectively than the original DC problem \cite{ho_dca-based_2021,tuy_dc_2016,lipp_variations_2015}. This paper uses the convex-concave procedure (CCP) proposed by Yuille and Rangarajan \cite{yuille_concave-convex_2003} for solving DC optimization problems given by \eqref{eq:dccp}. 
 
%The DCA model is based on fitting DC functions applied to unconstrained optimization problems. An advantage of $\ell$-DEP over the DCA model is that we have reduced the constraint number by half when transforming the unconstrained problem into a constrained problem.

\subsection{Convex-Concave Procedure}

The convex-concave procedure (CCP), also called concave-convex procedure \cite{yuille_concave-convex_2003}, is a majorization-minimization methodology that uses convex optimization tools to find a local optimum for DC problems through of a sequence of convex sub-problems. Briefly, the CCP method solves a constrained DC problem sequentially as follows: 
Given a feasible approximation $\alphav_k \in \mathds{R}^n$, the convex component $h_i:\mathds{R}^n \to \mathds{R}$ is approximated by 
\begin{equation} 
\tilde{h}_i(\alphav) = h_i(\alphav_k) + \lip \betav_i,\alphav-\alphav_k\rip, \quad \forall i=0,1,\ldots,m,
\end{equation}
where $\betav_i$ is a subgradient of $h_i$ at $\alphav_k$. In particular, $\betav_i = \nabla h_i(\alphav_k)$ if the function $h_i$ is differentiable at $\alphav_k$. Replacing $h_i$ by $\tilde{h}_i$ results the convex optimization problem
\begin{equation} \label{eq:convex-CPP}
    \begin{cases}
    \mathop{\text{minimize}}_{\alphav \in \mathds{R}^n} & \;  g_0(\alphav)- h_0(\alphav_k) - \lip \betav_0,\alphav-\alphav_k \rip\\ %+ t_k \sum_{i=1}^m s_i \\ 
    \text{subject to}  & \; g_i(\alphav)-h_i(\alphav_k)-\lip \betav_i,\alphav-\alphav_k \rip \le 0, \; i =1,\ldots.
    % & \; s_i \geq 0, \; i =1,\ldots,m, 
\end{cases}
\end{equation}
The solution of \eqref{eq:convex-CPP} is a new feasible approximation $\alphav_{k+1}$, and the process is repeated. The sequence $\{\alphav_k\}$ obtained solving \eqref{eq:convex-CPP} recursively is convergent and $\{f(\alphav_k)\}$ is non-increasing. Further details on the convergence of the sequence $\{\alphav_k\}_{k \geq 0}$ can found at \cite{lipp_variations_2015}. Algorithm \ref{alg:lCCP} summarizes the CCP method.

\begin{algorithm}[t]
\KwInput{Convex functions: $g_0,\ldots,g_m$ and $f_0,\ldots,f_m$.}
\KwOutput{$\alphav^*$ (Solution)}
\textbf{Initialize:} a feasible $\alphav_0\in \mathds{R}^n$ and $k=0$.\\
    \Repeat{converge}{
        Compute a subgradient $\betav_i$ of $h_i$ at $\alphav_k$, for all $i=0,\ldots,m$. \\
        Solve the convex problem:\\
        % $\vspace{-.3cm}\quad $ 
        $\begin{cases}
        \alphav_{k+1} = \mathop{\mbox{argmin}}_{\alphav \in \mathds{R}^n} & g_0(\alphav) - h_0(\alphav_k) - \lip\betav_0,\alphav-\alphav_k\rip\\% + t_k\sum_{i=1}^ms_i\\ 
          \qquad \qquad \mbox{s.t.} & g_i(\alphav) \le h_i(\alphav_k) + \lip\betav_i,\alphav-\alphav_k\rip, \quad i=1,\ldots,m.
        %   & s_i\ge0, \quad \forall i=1,\ldots,m.
        \end{cases}$\\
        % $t_{k+1}=\min\{\mu t_k,t_{max}\}$\\
        $k = k+1$}
	\Return{$\alphav^* = \alphav_{k}$}
\caption{\textsc{CCP}}
\label{alg:CCP}
\end{algorithm}

\section{Linear Dilation-Erosion Regressor}\label{sec:lder}

Predictive classification models categorize information based on a set of historical data. Predictive regression models are used to solve curve-fitting problems whose goal is to find a function that best fits a given data set. The adjusted mapping can be used for forecasting or predictions outside the data set. 

Recently, \cite{oliveira_linear_2021} introduced the linear dilation-erosion perceptron ($\ell$-DEP) for classification tasks. A linear dilation-erosion perceptron is given by a convex combination of the composition of linear transformations and two elementary operators from mathematical morphology \cite{ritter96c,araujo_class_2011}. Let us review the main concepts from mathematical morphology and the $\ell$-DEP model. We will subsequently present the linear dilation-erosion regressor, the predictive model of the regression type corresponding to the $\ell$-DEP regressor.

Mathematical morphology is mainly concerned with non-linear operators defined on complete lattices \cite{heijmans_mathematical_1995,soille_morphological_1999}. Complete lattices are partially ordered sets with well-defined supremum and infimum operations \cite{birkhoff_lattice_1993}. Dilations and erosions are the elementary operators from mathematical morphology. Given complete lattices $\mathbb{L}$ and $\mathbb{M}$, a dilation $\delta:\mathbb{L} \to \mathbb{M}$ and an erosion ${\varepsilon:\mathbb{L} \to \mathbb{M}}$ are operators such that 
\begin{equation}
\delta\left(\sup X\right)= \sup \{\delta(\bx):\bx \in X\} \qeq {\varepsilon\left(\inf X\right)= \inf \{\varepsilon(\bx):\bx \in X\}},
\end{equation} 
for all $X \subseteq \mathbb{L}$ \cite{heijmans_mathematical_1995}. For example, consider the complete lattices $\mathbb{L} = \bar{\mathds{R}}^n$ and $\mathbb{M} = \bar{\mathds{R}}$, where $\bar{\mathds{R}} = \mathds{R}\cup \{-\infty,+\infty\}$. Given vectors $\ba,\bb \in \mathds{R}^n$, the operators $\delta_{\ba},\varepsilon_{\bb}:\bar{\mathds{R}}^n \to \bar{\mathds{R}}$ defined by
\begin{equation}  \label{eq:erodil}
    \delta_{\ba}(\bx)=\max_{j=1:n}\{x_j + a_j\} \quad \mbox{and} \quad  \varepsilon_{\bb}(\bx)=\min_{j=1:n} \{x_j - b_j\}, \quad \forall \bx \in \bar{\mathds{R}}^n,
\end{equation}  
are respectively a dilation and an erosion \cite{sussner_morphological_2011}. Note that dilations and erosions given by \eqref{eq:erodil} satisfy the duality identity 
% \begin{equation} \label{eq:duality}
$\varepsilon_{\bb}(-\bx) = - \delta_{\bb}(\bx)$, for all $\bx \in \bR^n$. 
% \end{equation}

A dilation-erosion perceptron (DEP) is given by a convex combination of a dilation and an erosion defined by \eqref{eq:erodil}. The reduced dilation-erosion perceptron ($r$-DEP) proposed by Valle is an improved version of the DEP model obtained using concepts from vector-valued mathematical morphology \cite{valle_reduced_2020}. The $\ell$-DEP model is a particular but powerful $r$-DEP classifier \cite{oliveira_linear_2021}. Formally, given a one-to-one mapping $\sigma$ from the set of binary class labels $\mathbb{C}$ to $\{+1,-1\}$, an $\ell$-DEP classifier is defined by the equation $y=\sigma^{-1}f\tau^{\ell}(\bx)$, where $f:\mathds{R}\to \{-1,+1\}$ is a threshold function and $\tau^{\ell}:\mathds{R}^n \to \mathds{R}$ is the decision function given by
  \begin{align}\label{eq:tau-DC}
        \tau^{\ell}(\bx) &= \delta_{\ba}(W\bx) - \delta_{\bb}(M\bx),
\end{align}
where $W \in \mathds{R}^{r_1 \times n}$, $M \in \mathds{R}^{r_2 \times n}$, $\ba \in \mathds{R}^{r_1}$, and $\bb \in \mathds{R}^{r_2}$. We would like to point out that  the erosion has been replaced by minus a dilation in \eqref{eq:tau-DC} using the duality identity. Equivalently, the decision function $\tau^{\ell}$ satisfies
\begin{align} \label{eq:tau-DC2}
        \tau^{\ell}(\bx) &= \max_{i=1:r_1}\{\lip \bw_i,\bx\rip+ a_i\} -  \max_{j=1:r_2}\{\lip\bm_j,\bx\rip+ b_j\},
\end{align}     
where $\mathbf{a} = (a_1,\ldots,a_{r_1}) \in \mathds{R}^{r_1}$ and $\mathbf{b} = (b_1,\ldots,b_{r_2}) \in \mathds{R}^{r_2}$, and $\bw_i \in \mathds{R}^n$ and $\bm_i \in \mathds{R}^n$ are rows of $W \in \mathds{R}^{r_1 \times n}$ and $M \in \mathds{R}^{r_2 \times n}$, respectively. From the last identity, we can identify $\tau^{\ell}$ with a piece-wise linear function \cite{wang_general_2004}. Moreover, from Theorem 4.3 in \cite{goodfellow_maxout_2013}, the decision function $\tau^{\ell}$ is an universal approximator, i.e., it is able to approximate any continuous-valued function from a compact set on $\mathds{R}^n$ to $\mathds{R}$ \cite{goodfellow_maxout_2013,stone_generalized_1948}.
Consequently, an $\ell$-DEP model can theoretically solve any binary classification problem. In addition, the decision function $\tau^{\ell}$ can be identified with a maxout network with two hidden units \cite{goodfellow_maxout_2013}.

Because the decision function of an $\ell$-DEP model is a universal approximator, $\tau^\ell$ given by \eqref{eq:tau-DC} can also be used as a predictive model for regression tasks. In other words, it is possible to use the mapping $\tau^{\ell}$ as the prediction function that maps a set of independent variables in $\mathds{R}^n$ to a dependent variable in $\mathds{R}$. In this case, we refer to the operator $\tau^{\ell}:\mathds{R}^n \to \mathds{R}$ given by \eqref{eq:tau-DC} as a linear dilation-erosion regressor {($\ell$-DER)}. In this paper, the parameters $(\bw_i^T,a_i)\in\mathds{R}^{n+1}$ and $(\bm_j^T,b_j)\in \mathds{R}^{n+1}$, for $i=1,\ldots,r_1$ and $j=1,\ldots,r_2$, are determined by minimizing the squares of the difference between the predicted and desired values. The following section addresses an approach for training an $\ell$-DER model. 

\section{Training the $\ell$-DER using CCP
}\label{sec:approaches}

In this section, we present approaches for training an $\ell$-DER model using a set ${\mathcal{T}=\{(\bx_i, y_i):i=1:m\}}\subset\mathds{R}^n \times \mathds{R}$, called training set. The goal is to find the parameters of an $\ell$-DER model such that the estimate $\tau^{\ell} (\bx_i)$ approaches the desired output $y_i$ according to some loss function. Recall that the parameters of an $\ell$-DEP regressor are the matrices $W \in \mathds{R}^{r_1 \times n}$ and $M \in \mathds{R}^{r_2 \times n}$ as well as the vectors $\mathbf{a} = (a_1,\ldots,a_{r_1}) \in \mathds{R}^{r_1}$ and $\mathbf{b} = (b_1,\ldots,b_{r_2}) \in \mathds{R}^{r_2}$. To simplify the exposition, the parameters of an $\ell$-DER are arranged in a vector
\begin{equation} \label{eq:alphav}
\alphav = (\bw_1,a_1,...,\bw_{r_1},a_{r_1},\bm_1,b_1,...,\bm_{r_2},b_{r_2})\in\mathds{R}^{(r_1+r_2)(n+1)},
\end{equation} 
where $\bw_i$ and $\bm_i$ denote rows of $W \in \mathds{R}^{r_1 \times n}$ and $M \in \mathds{R}^{r_2 \times n}$, respectively. During the training, an $\ell$-DER is interpreted as a function of its parameters, that is, $\tau^\ell(\bx) \equiv \tau^\ell(\bx;\alphav)$.

In this paper, the mean squared error (MSE) defined as follows using the training set ${\mathcal{T}=\{(\bx_i, y_i):i=1,\ldots,m\}}\subset\mathds{R}^n \times \mathds{R}$ is considered as the loss function: 
\begin{equation} \label{eq:MSE}
\mathtt{MSE}(\mathcal{T},\alphav)= \frac{1}{m}\sum_{i=1}^m (y_i-\tau^{\ell}(\bx_i;\alphav))^2,
\end{equation}
As a consequence, the parameters of the $\ell$-DER are determined by solving the optimization problem
\begin{equation}\label{eq:mse}
\mathop{\mbox{minimize}}_{\alphav} \frac{1}{m}\sum_{i=1}^m (y_i-\tau^{\ell}(\bx_i;\alphav))^2.
\end{equation}

\subsection{Training Based on the Convex-Concave Programming}

Inspired by the methodology developed by Charisopoulos and Maragos for training morphological perceptrons \cite{charisopoulos_morphological_2017}, we  reformulate the unrestricted optimization problem \eqref{eq:mse} as a constrained DC problem. Precisely, by setting $\xi_i = y_i-\tau^{\ell}(\bx_i)$, the unrestricted problem \eqref{eq:mse} corresponds to
\begin{equation}
\begin{cases}
      \mathop{\mbox{minimize}}_{W,\ba,M,\bb,\bxi} & \frac{1}{m} \sum_{i=1}^m \xi_i^2 \\ 
    \mbox{subject to} & \tau^{\ell}(\bx_i) = y_i-\xi_i, \quad i=1,...,m. 
\end{cases}    
\end{equation}
In other words, the $\ell$-DER can be trained by solving the following problem
\begin{equation} \label{eq:DCCP}
    \begin{cases}
    \mathop{\mbox{minimize}}_{W,\ba,M,\bb,\bxi}             & \frac{1}{m} \sum_{i=1}^m \xi_i^2 \\ 
    \mbox{subject to} & \delta_{\ba}(W\bx_i) + \xi_i=  \delta_{\bb}(M\bx_i) + y_i, \quad i=1,...,m.
    \end{cases}
\end{equation}
Note that the objective function in \eqref{eq:DCCP} is a convex quadratic function. Moreover, the functions at both sides of the equality constraints are convex. Thus, we can view the constraints as DC functions, and the optimization problem \eqref{eq:DCCP} can be solved using  Algorithm \ref{alg:CCP} by dealing appropriately with the equality constraints. 

We first transform each of the equality constraints into two inequality constraints:
\begin{equation} \label{eq:2inequalities}
\delta_{\ba}(W\bx_i) + \xi_i \leq \delta_{\bb}(M\bx_i) + y_i
\quad \text{and} \quad
\delta_{\bb}(M\bx_i) + y_i \leq \delta_{\ba}(W\bx_i) + \xi_i, 
\end{equation}
for all $i=1,\ldots,m$. Because a majorant of a set is also a majorant of each of its elements and recalling that
\begin{equation}
    \delta_{\ba}(W\bx_i) = \max_{j=1:r_1} \{\lip \bw_j,\bx_i\rip + a_j\} \qeq \delta_{\bb}(M\bx_i) = \max_{j=1:r_2} \{\lip \bm_j,\bx_i\rip + b_j\},
\end{equation} the two inequalities in \eqref{eq:2inequalities} yield 
\begin{align}
\lip \bw_l,\bx_i \rip + a_l +\xi_i \leq \delta_{\bb}(M\bx_i) + y_i, \quad i \in \mathcal{I}, l \in \mathcal{L}_1, \\
\lip \bm_l,\bx_i \rip + b_l + y_i \leq \delta_{\ba}(W\bx_i) + \xi_i, \quad i \in \mathcal{I}, l \in \mathcal{L}_2,
\end{align}
where $\mathcal{I} = \{1,\ldots,m\}$, $\mathcal{L}_1=\{1,\ldots,r_1\}$, and $\mathcal{L}_2 = \{1,\ldots,r_2\}$. Thus, \eqref{eq:DCCP} can be equivalently written as 
\begin{equation} \label{eq:DCCP-2}
\begin{cases}
    \mathop{\mbox{minimize}}_{W,\ba,M,\bb,\bxi}             & \frac{1}{m} \sum_{i=1}^m \xi_i^2 \\ 
    \mbox{subject to} & \big(\lip \bw_l,\bx_i \rip + a_l +\xi_i -y_i\big) - \delta_{\bb}(M\bx_i) \leq 0, \; i \in \mathcal{I}, l \in \mathcal{L}_1, \\ &
\big(\lip \bm_l,\bx_i \rip + b_l -\xi_i + y_i\big) - \delta_{\ba}(W\bx_i) \leq 0, \; i \in \mathcal{I}, l \in \mathcal{L}_2.
    \end{cases}
\end{equation}
Note that  \eqref{eq:DCCP-2} can be identified with a DC optimization problem \eqref{eq:dccp}. Furthermore, the convex functions
\begin{equation} 
    h_{i1}(\alphav) = \delta_{\bb}(M\vetx_i) \qeq 
    h_{i2}(\alphav) = \delta_{\ba}(M\vetx_i),
\end{equation}
can be approximated by the affine functions
\begin{equation}
    \tilde{h}_{i1}(\alphav) = \lip \bm_{j_{i1}},\bx_i\rip + b_j  \qeq 
    \tilde{h}_{i2}(\alphav) = \lip \bw_{j_{i2}},\bx_i\rip + a_j,
\end{equation}
where 
\begin{equation} \label{eq:indexes}
        j_{i1}=\argmax_{j=1,\ldots,r_2}\{\lip \bm_j,\bx_i \rip+ b_j\} \qeq 
        j_{i2}=\argmax_{j=1,\ldots,r_1}\{\lip \bw_j,\bx_{i} \rip + a_j\},
    \end{equation}
for all $i \in \mathcal{I}$.

% Defining the convex functions
% \begin{equation*}
%     \begin{array}{lcl}
%         g_0(\alphav,\bxi)=\frac{1}{m}\sum_{i=1}^m\xi_i^2  & \qquad & h_0(\alphav) = 0, \\
%         g_{il}(\alphav,\bxi)=\bw_l^T\bx_i + a_l+\xi_i,  & & h_{i1}(\alphav) = \max_{j=1:r_2}\{\bm_j^T\bx_i+b_j\}+y_i,  \\
%         g_{il}(\alphav,\bxi)=\bm_{l}^T\bx_{i} + b_{l}-\xi_{i}, & & h_{i2}(\alphav) = \max_{j=1:r_1}\{\bw_j^T\bx_i+a_j\}-y_i,
%     \end{array}
% \end{equation*}
% where $\alphav \in \mathds{R}^{(r_1+r_2)(n+1)}$ is given by \eqref{eq:alphav}, we can identify \eqref{eq:DCCP-2} with a DC optimization problem \eqref{eq:dccp}. 
%
% The subgradient of $h_{i1}$ and $h_{i2}$ at
% \begin{equation} \label{eq:alphak-DER}
% \alphav_k = (\bw_1^{(k)},a_1^{(k)},...,\bw_{r_1}^{(k)},a_{r_1}^{(k)},\bm_1^{(k)},b_1^{(k)},...,\bm_{r_2}^{(k)},b_{r_2}^{(k)})
% \end{equation}
% are given by
%     \begin{equation}\label{eq:betai}
%         \betav_{i1} = \bv^{i,r_1+j_{i1}} \qeq
%         \betav_{i2} = \bv^{i,j_{i2}}, \quad \forall i \in \mathcal{I},
%     \end{equation}
% where 
%     \begin{equation}
%         j_{i1}=\argmax_{j=1,\ldots,r_2}\{\lip \bm^{(k)}_j,\bx_i \rip+ b^{(k)}_j\} \qeq 
%         j_{i2}=\argmax_{j=1,\ldots,r_1}\{\lip \bw^{(k)}_j,\bx_{i} \rip + a^{(k)}_j\},
%     \end{equation}
% and ${\bv^{i,s} = (\bv^{i,s}_1,\dots,\bv^{i,s}_{r_1+r_2})\in\mathds{R}^{(r_1+r_2)(n+1)}}$ is such that
%     \begin{equation}
%     \bv^{i,s}_{j} = 
%     \begin{cases}
%         (\bx_i,1), & j=s,\\
%         ({\bf 0},0), & \text{otherwise}.
%     \end{cases}   
%     \end{equation}
%

Using the affine approximations of the convex components $h_{i1}$ and $h_{i2}$ of the constraints, we obtain the following quadratic problem which is solved at each iteration of the CCP method used for training an $\ell$-DER model: 
\begin{equation} 
\begin{cases}
       \mathop{\mbox{minimize}}_{W,\ba,M,\bb,\bxi,\bp,\bq}\quad & \frac{1}{m}\sum_{i=1}^m\xi_i ^2 \label{prob:lccp}\\
       \quad \mbox{subject to}\quad
       & \lip\bw_l-\bm_{j_{i1}},\bx_i\rip + a_l - b_{j_{i1}} \le y_i- \xi_i , \; i \in \mathcal{I},  l \in \mathcal{L}_1\\
       & \lip\bm_l-\bw_{j_{i2}},\bx_i\rip + b_l - a_{j_{i2}} \le \xi_i- y_i, \; i \in \mathcal{I},  l \in \mathcal{L}_2 \\
    %   & p_i,q_i\ge 0, \; \quad i \in \mathcal{I}.
\end{cases}
\end{equation}

The iterations are performed until a maximum number of iterations is reached or when the difference of the objective function at two consecutive iterations is less than a tolerance $\epsilon$. Algorithm \ref{alg:lCCP} summarizes the training of the $\ell$-DER model with the CCP method.

For our models, the starting point for the optimization problems was produced using a deterministic strategy presented in \cite{ho_dca_2020}.
% , that uses the KKZ method \cite{katsavounidis_new_1994}, Voronoi partitions \cite{moller_lectures_1994} and possible linear problems. 
Roughly speaking, we first use the KKZ method to find a subset of centroids of the training data \cite{katsavounidis_new_1994}. The training data is then grouped using Voronoi partitions. Finally, the starting point of the DC optimization method is obtained by (traditional) linear least squares data-fit on the Voronoi regions. 

\begin{algorithm}[h]
\KwInput{Training set $\mathcal{T} = \{(\bx_i,y_i):i \in \mathcal{I}\}$ and the parameters $k_{max}$ and $\epsilon$.}
\KwOutput{$\{W,\ba,M,\bb$\}}
\textbf{Initialize:} $W$, $\ba$, $M$, $\bb$, $\bxi \text{ and } k=0$\\
    \Repeat{$k \ge k_{max}$ or {\em the difference in the objective function is less than an} $\epsilon$ }{
    Determine the indexes $j_{i1}$ and $j_{i2}$ using \eqref{eq:indexes} \\
 Compute $W$, $\ba$, $M$, $\bb$, $\bxi$ for $i=1,\ldots,m$ solving \eqref{prob:lccp}\\
%  $\lambda_{k+1}=\min\{\mu \lambda_k,\lambda_{max}\}$\\
 $k = k+1$}
	\Return{$W,\ba,M,\bb$}
\caption{{\textsc{$\ell$-CCP}}}
\label{alg:lCCP}
\end{algorithm}

\section{Computational Experiments}\label{sec:experiments}

This section presents some computational experiments to evaluate the performance of the proposed $\ell$-DER model trained using the CCP method for regression tasks. Furthermore, we compare the $\ell$-DER model with other models from the literature. Namely, we consider the multilayer perceptron (MLP) \cite{haykin_neural_2008}, the support vector regressor (SVR) \cite{smola02}, the hybrid linear/morphological extreme learning machine (HLM-ELM) \cite{sussner_extreme_2020}, the morphological dense network (MDN) \cite{mondal_morphological_2020}, and the maxout network \cite{goodfellow_maxout_2013}. Because the $\ell$-DER yields a continuous piecewise linear function, we also compare it with the regressor trained by the difference of convex algorithm (DCA) proposed by Ho et al.  \cite{ho_dca-based_2021}.

    \begin{table}[h]
        \centering
        \caption{Hyper-parameters for the datasets.}
        \def\arraystretch{1.2} %\tabcolsep=10pt
        % \resizebox{\textwidth}{!}{
        \scriptsize
        \begin{tabular}{|c||c|ccc|c|}
         \hline
         {\textbf{Model}}  & {\textbf{Parameters}} \\
        \hline
        \textbf{MLP} & Default \texttt{Sklearn}\\
        \hline
        \textbf{SVR} & Default \texttt{Sklearn}\\
        \hline
        \textbf{HLM-ELM} & linear neurons=132\\
                 & morfological neurons=141\\
        \hline
        \textbf{MDN} & $(r_1^1,r_2^1,\text{out}^1)=(100,100,100 )$\\
            & $(r_1^2,r_2^2,\text{out}^2)=(100,100,1)$ \\
         \hline
        \textbf{MAXOUT} & $(r_1,r_2)=(3,3)$ \\
        \hline
        \textbf{DCA} & $(r_1,r_2)=(3,2)$\\
        \hline
      \textbf{ $\ell$-DER} & $(r_1,r_2)=(3,2)$\\
         \hline
        \end{tabular}
        % }   
        \label{tab:melhoresparam}
    \end{table}

% Let us now evaluate the performance of the proposed $\ell$-DER model on regression datasets from the Penn Machine Learning Benchmarks (PMLB), a significant benchmark suite for machine learning evaluation and comparison \cite{olson_pmlb_2017}. The chosen datasets have small or medium sizes. According to \autoref{tab:melhoresparam}, we fixed the hyper-parameters $r_1 = r_2 = 7$ of the {$\ell$-DER} models for all datasets. Other models' hyper-parameters can be found in the last column of \autoref{tab:melhoresparam}.

% We compared the performance of the $\ell$-DER model trained using the approaches described in the previous section. 
% % The $\ell$-DER trained using the SGD method has implemented using the \texttt{TensorFlow API} (TF). Precisely, because $\ell$-DER is equivalent to a maxout network \cite{goodfellow_maxout_2013}, we used \texttt{tensorflow-addons}, which provides extra functionalities and include the maxout layers. 
% For the two approaches, we used the \texttt{CVXPY} package \cite{diamond_cvxpy_2016} with the \texttt{MOSEK} solver \cite{aps_mosek_2020}. 
% % For the pCCP problem, in particular, we used the DCCP extension for the \texttt{CVXPY} available at \url{https://github.com/cvxgrp/dccp}.

% We would like to point out that we handled missing data using sklearn's \texttt{SimpleImputer()} command. Furthermore, we partitioned the data set into training and test sets using the sklearn's \texttt{KFold()} command with $k=5$. Finally, we used the regularization coefficient to measure the performance of the regressor $\ell$-DER for each of the training approaches. 

We evaluated the performance of the proposed $\ell$-DER and the other models from the literature using regression datasets from the Penn Machine Learning Benchmarks (PMLB), a significant benchmark suite for machine learning evaluation and comparison \cite{olson_pmlb_2017}. The chosen datasets are listed in \autoref{tab:dset_openml}. Note that we considered small and medium-sized datasets. We would like to point out that we handled possible missing data using \texttt{sklearn}'s \texttt{SimpleImputer} command. Furthermore, we partitioned the data set into training and test sets using the \texttt{sklearn}'s \texttt{KFold}, with the default number of splits (\texttt{n\_splits=5}).  

We trained the $\ell$-DER model using the \texttt{CVXPY} package \cite{diamond_cvxpy_2016} with the \texttt{MOSEK} solver \cite{aps_mosek_2020}. Because the $\ell$-DER is equivalent to a continuous piecewise linear function, we adopt the same hyperparameters $r_1 =3$ and $r_2 = 2$ as Ho et al. for all datasets \cite{ho_dca-based_2021}. For a fair comparision between the continuous piecewise linear models, we consider $r_1=r_2 = 3$ for the  maxout network \cite{goodfellow_maxout_2013}. Finally, the other models' hyperparameters are set to the default value of the \texttt{sklearn} or as far as possible to values reported in the literature \cite{sklearn_api,mondal_morphological_2020,sussner_extreme_2020}. \autoref{tab:melhoresparam} summarizes the hyperparameters used in our computational experiment. 

\begin{figure}
    \centering
    \includegraphics[width=.6\columnwidth]{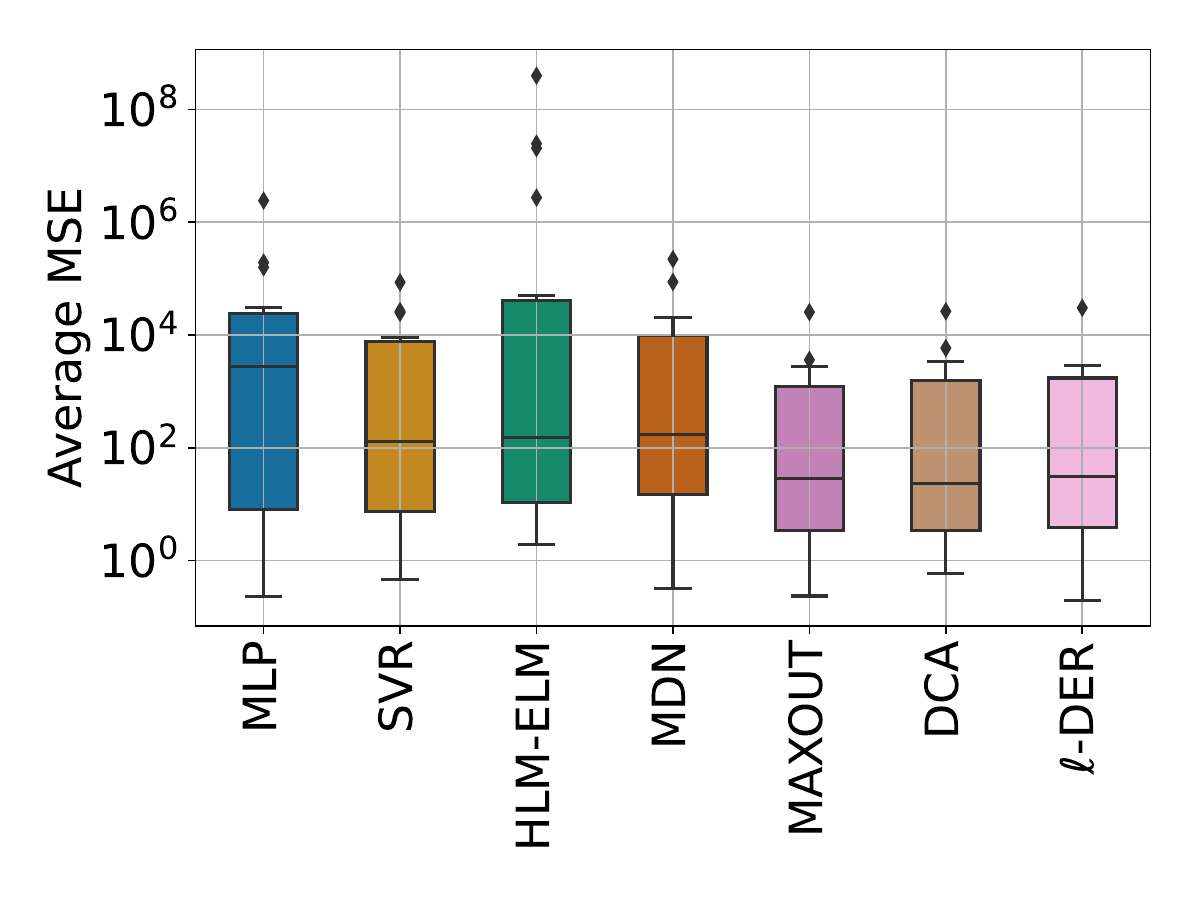}
    \includegraphics[width=.6\columnwidth]{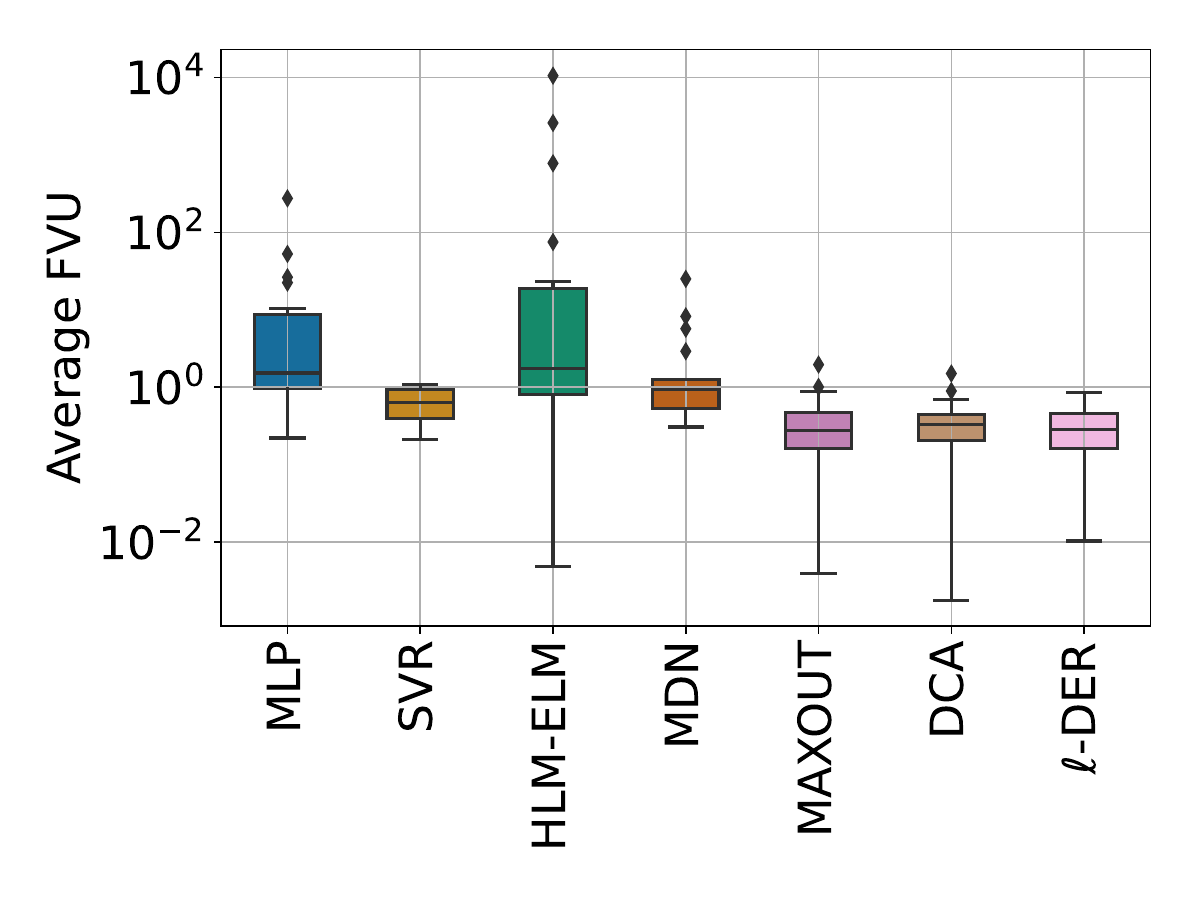}
    \includegraphics[width=.6\columnwidth]{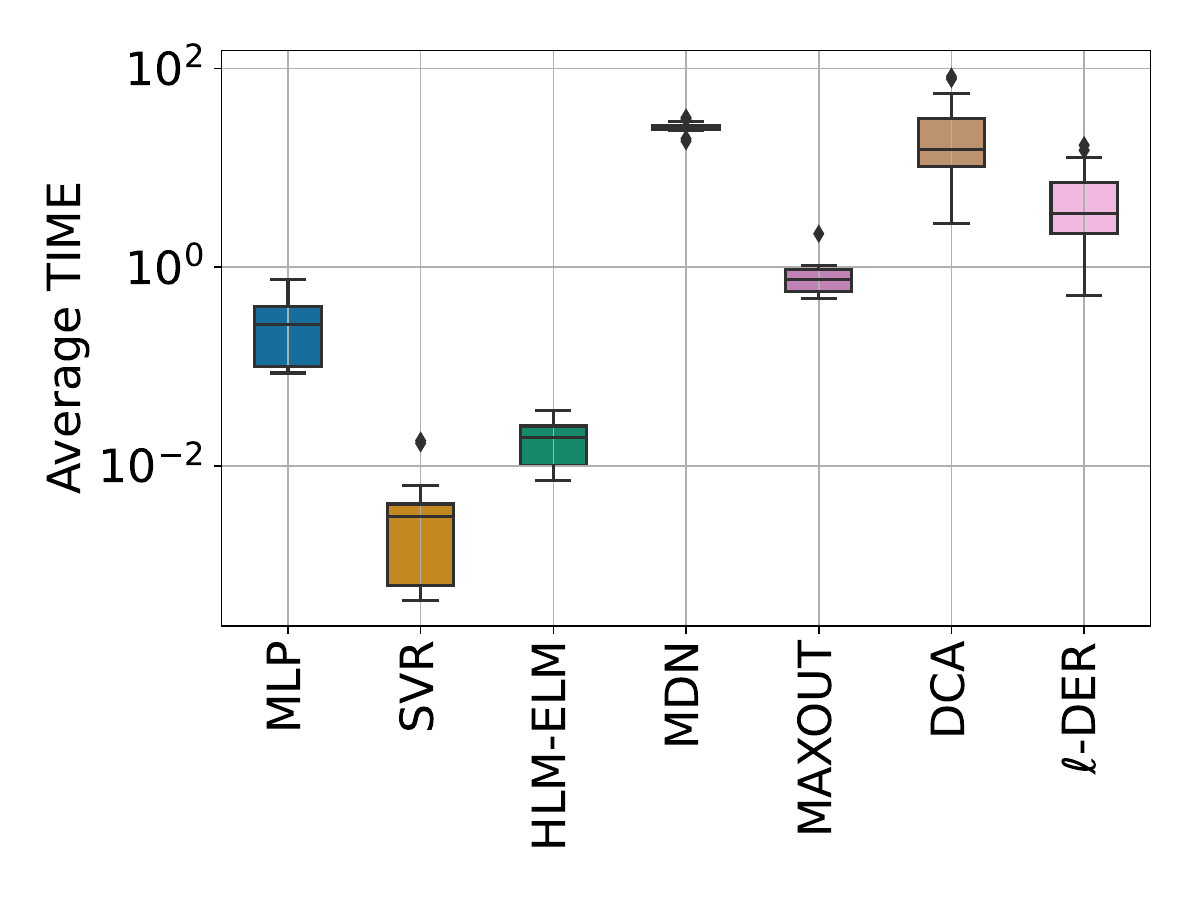}
    \caption{Boxplots of the MSE and FVU scores and average time required to train the machine learning regressors.} 
    \label{fig:box_scores_times}
\end{figure}

We evaluated the performance of the regression models quantitatively using the well-known mean squared error (MSE) and the fraction of variance unexplained (FVU) given by the following equations:
\begin{equation}
\displaystyle
\mathtt{MSE} = \frac{1}{m} \sum_{i=1}^m (y_i - f(\bx_i))^2 \qeq
\mathtt{FVU} = \frac{\sum_{i=1}^m (y_i - f(\bx_i))^2}{\sum_{i=1}^m (y_i - \bar{y})^2},     
\end{equation}
where $f:\mathds{R}^n \to \mathds{R}$ denotes the regressor and $\bar{y} = \frac{1}{m}\sum_{i=1}^m y_i$ is the average of the output values. Note that the FVU is the ratio between the MSEs produced by the regressor and a dummy model, which consistently predicts the average output values. The less the MSE and the FVU scores, the better is the regressor $f$. \autoref{tab:score} contains the average and the standard deviation of the FVU score obtained from the regressors using $5$-fold cross-validation on the considered datasets. The best outcome for each dataset has been typed using boldface numbers. The boxplots shown in \autoref{fig:box_scores_times} summarize the MSE and FVU scores and the execution time (in seconds) taken for training the regressors in the sixteen datasets.

From the boxplot shown in \autoref{fig:box_scores_times}, the $\ell$-DER trained with the CCP procedure achieved the best performance (concerning both MSE and FVU) together with the maxout network and the continuous piecewise linear regressor trained using the DCA algorithm. The HLM-ELM and MLP networks presented the worst performance according to both MSE and FVU scores. 

\begin{figure}
    \centering
    \begin{tabular}{cc}
    a) Mean square error (MSE) & b) Fraction of variance unexplained (FVU) \\
    \includegraphics[width=.4\columnwidth]{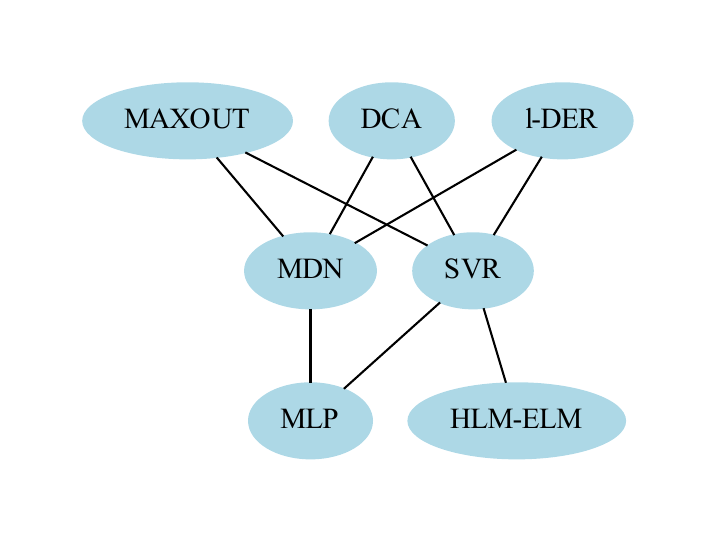} &
    \includegraphics[width=.3\columnwidth]{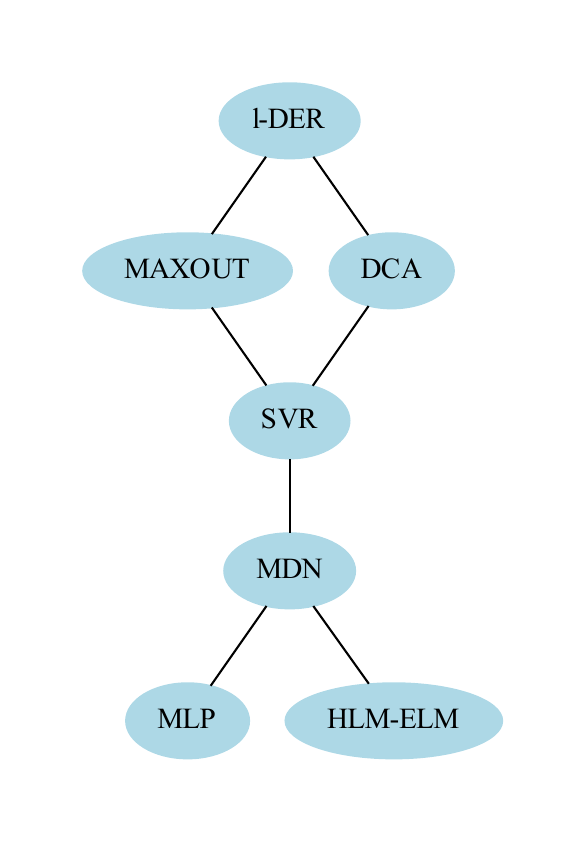}
    \end{tabular}
    \caption{Hasse diagram of Wilcoxon hypothesis test using the MSE and FVU scores.} 
    \label{fig:hasse_diagram}
\end{figure}

The Hasse diagrams depicted in \autoref{fig:hasse_diagram} illustrate the outcome of Wilcoxon's nonparametric statistical test with a confidence level of $95\%$ for the MSE and FVU scores \cite{weise15}. In the Hasse diagram depicted in \autoref{fig:hasse_diagram}, an edge means that the model on top statistically outperformed the one below. Thus, the topmost models are the best-performing models. Moreover, transitivity is valid in the Hasse diagram. Note from \autoref{fig:hasse_diagram} that the $\ell$-DER achieved the best performance in terms of FVU score. The maxout network and the model trained using DCA are competitive but superior to the remaining models for the FVU. In terms of the MSE score, the $\ell$-DER, maxout network, and the model trained using DCA are all competitive and superior to the other models in these regression tasks. 

% Although the continuous piecewise linear model trained with the DCA outperformed the $\ell$-DER concerning the FVU score, they achieved comparable performance for the FVU score. Moreover, t
Finally, note that training the $\ell$-DER model using the CCP method is, on average, faster than the DCA model. Indeed, the $\ell$-DER model's processing time is approximately 21\% of the time the DCA model spends on training. The slowness of DCA probably follows because it requires solving more complex optimization problems than the ones obtained using CCP.

% The regularization coefficient, also known as $R^2$ score, is a statistical measure that defines how close the data are to the fitted regression hypersurface. The closer $R^2$ is to 1, the better the model fits the datasets. The $R^2$ metric can be applied through another metric called the MARE score (mean absolute relative error). The MARE is defined as the ratio between the MSE of the regressor to be evaluated and the MSE of a base regressor, which cannot have the null MSE\footnote{If the MSE of the base regressor is null, then each term of the sum that sets the MSE is zero. Then the base model fits the data.}. In this case, we have $R^2= 1-$MARE.

% In the \autoref{tab:score} contains the mean and the standard deviation of the $R^2$ score obtained from the regressors using $5$-fold cross-validation. The boxplots shown in  \autoref{fig:box_scores_times} summarizes the normalized scores depicted on this table as well as the execution time (in seconds) taken for training the regressors.

\begin{table}[t] 
    \centering
    \caption{Information about the datasets of PMLB for regression tasks.}
\resizebox{\textwidth}{!}{
    \begin{tabular}{clccc}
       \textbf{Dataset} & \textbf{Name} 			& \textbf{Instances} & \textbf{Feature} & \textbf{PMLB Code}\\
       \hline
        \textbf{D1} & ANALCATDATA\_VEHICLE & 48 & 4 & 485\_analcatdata\_vehicle\\
        \textbf{D2} & BODYFAT & 252 & 14 & 560\_bodyfat\\
        \textbf{D3} & CHSCASE\_GEYSER1 & 222 & 2 & 712\_chscase\_geyser1\\
        \textbf{D4} & CLOUD & 108 & 5 & 210\_cloud\\
        \textbf{D5} & CPU & 209 & 7 & 561\_cpu\\
        \textbf{D6} & ELUSAGE & 55 & 2 & 228\_elusage\\
        \textbf{D7} & MACHINE\_CPU & 209 & 6 & 230\_machine\_cpu\\
        \textbf{D8} & PM10 & 500 & 7 & 522\_pm10\\
        \textbf{D9} & PWLINEAR & 200 & 10 & 229\_pwLinear\\
        \textbf{D10} & RABE\_266 & 120 & 2 & 663\_rabe\_266\\
        \textbf{D11} & RMFTSA\_LADATA & 508 & 10 & 666\_rmftsa\_ladata\\
        \textbf{D12} & SLEUTH\_CASE1202 & 93 & 6 & 706\_sleuth\_case1202\\
        \textbf{D13} & SLEUTH\_EX1605 & 62 & 5 & 687\_sleuth\_ex1605\\
        \textbf{D14} & VINEYARD & 52 & 2 & 192\_vineyard\\
        \textbf{D15} & VINNIE & 380 & 2 & 519\_vinnie\\
        \textbf{D16} & VISUALIZING\_ENVIRONMENTAL & 111 & 3 & 678\_visualizing\_environmental\\
        \textbf{D17} & VISUALIZING\_GALAXY & 323 & 4 & 690\_visualizing\_galaxy\\
\bottomrule
    \end{tabular}
    }\label{tab:dset_openml}
\end{table}

% \begin{figure}
%     \centering
%     % \includegraphics[width=.49\columnwidth]{BoxPlot_Datasets R2.pdf}
%     % \includegraphics[width=.49\columnwidth]{BoxPlot_Datasets TIME.pdf}
%     \includegraphics[width=.9\columnwidth]{BoxPlot_R2.pdf}
%     \includegraphics[width=.9\columnwidth]{BoxPlot_Time.pdf}
%     % \includegraphics[width=.49\columnwidth]{HoTDiagram_R2.pdf}
%     \caption{ Boxplots of the $R^2$ score and average time required to train the $\ell$-DEP regressors.} 
%     \label{fig:box_scores_times}
% \end{figure}

% \begin{figure}[h]
%     \centering
%     \includegraphics[width=.7\columnwidth]{BoxPlot_Time.pdf}
%     \caption{Average time required to train the machine learning regressors.} 
%     \label{fig:box_time}
% \end{figure}

 %HoT

\section{Concluding Remarks}\label{sec:concluding}

This paper introduced the linear dilation-erosion regressor ($\ell$-DER), which is given by a convex combination of the composition of linear transformations and elementary morphological operators. Precisely, an $\ell$-DER is defined by the DC function $\tau^{\ell}$ given by \eqref{eq:tau-DC}. Because $\tau^{\ell}$ is a continuous piecewise linear function, an $\ell$-DER is a universal approximator \cite{wang_general_2004}.   

An $\ell$-DER model can be trained by minimizing the mean square error given by \eqref{eq:mse} using a training set ${\mathcal{T}=\{(\bx_i, y_i):i=1:m\}}\subset\mathds{R}^n \times \mathds{R}$. This paper proposed to train an $\ell$-DER $\tau^{\ell}$ using DC optimization. The approach solves a constrained DC problem using the CCP method \cite{lipp_variations_2015}.

We compared the $\ell$-DER trained using CCP with other approaches from the literature using several regression tasks. According to the preliminary computational experiments, the $\ell$-DER trained using CCP is competitive with state-of-the-art methods like the SVR. Furthermore, regarding the computational complexity of training the $\ell$-DER, solving the quadratic optimization subproblems incurs the highest cost.

In the future, we plan to investigate further the performance of the $\ell$-DER model trained by the CCP method. In particular, we intend to study the effects of the hyper-parameters -- such as the shape of the matrices $W$ and $M$ -- on the curve fitting capability of the $\ell$-DER. 

    \begin{table}[p]
    \begin{center}
    \caption{Average and standard deviation of the FVU scores on datasets in \autoref{tab:dset_openml}.}
    \resizebox{!}{\textheight}{
    \rotatebox{90}{
    \def\arraystretch{1.5} \tabcolsep=8pt
\begin{tabular}{lccccccc}
\toprule
\textbf{Dataset} &     \textbf{MLP} &              \textbf{SVR} &                  \textbf{HLM-ELM} &               \textbf{MDN} &           \textbf{MAXOUT} &              \textbf{DCA} &             \textbf{$\ell$-DER} \\
\midrule
\textbf{D1}      &     2.490 $\pm$ 0.233 &  1.091 $\pm$ 0.087 &          0.656 $\pm$ 0.394 &   1.154 $\pm$ 0.242 &  0.299 $\pm$ 0.199 &  \textbf{0.315 $\pm$ 0.185} &  0.379 $\pm$ 0.183 \\
\textbf{D2}      &     1.277 $\pm$ 0.414 &  0.209 $\pm$ 0.058 &          0.225 $\pm$ 0.103 &   0.904 $\pm$ 1.010 &  0.043 $\pm$ 0.030 &  0.028 $\pm$ 0.029 & \textbf{ 0.027 $\pm$ 0.037} \\
\textbf{D3}      &    2.6e1 $\pm$ 4.539 &  0.265 $\pm$ 0.016 &          2.3e1 $\pm$ 2.4e1 &   1.228 $\pm$ 0.541 &  0.251 $\pm$ 0.050 &  0.236 $\pm$ 0.032 &  \textbf{0.234 $\pm$ 0.034} \\
\textbf{D4}      &     0.235 $\pm$ 0.182 &  0.361 $\pm$ 0.174 &          1.724 $\pm$ 1.005 &   0.307 $\pm$ 0.124 &  0.247 $\pm$ 0.133 &  0.529 $\pm$ 0.532 & \textbf{ 0.202 $\pm$ 0.108} \\
\textbf{D5}      &     1.215 $\pm$ 0.200 &  1.013 $\pm$ 0.060 &          7.5e1 $\pm$ 1.1e2 &   0.556 $\pm$ 0.227 &  \textbf{0.037 $\pm$ 0.022} &  0.166 $\pm$ 0.300 &  0.099 $\pm$ 0.062 \\
\textbf{D6}      &     3.506 $\pm$ 0.883 &  0.703 $\pm$ 0.233 &          1.896 $\pm$ 0.754 &   \textbf{0.340 $\pm$ 0.175} &  0.409 $\pm$ 0.259 &  0.367 $\pm$ 0.131 &  0.344 $\pm$ 0.133 \\
\textbf{D7}      &     1.255 $\pm$ 0.174 &  1.017 $\pm$ 0.063 &          1.1e4 $\pm$ 1.7e4 &   0.511 $\pm$ 0.191 &  \textbf{0.129 $\pm$ 0.064} &  0.229 $\pm$ 0.129 &  0.148 $\pm$ 0.090 \\
\textbf{D8}      &     0.868 $\pm$ 0.053 &  \textbf{0.724 $\pm$ 0.033} & 2.544 $\pm$ 0.572 &   0.977 $\pm$ 0.112 &  0.886 $\pm$ 0.056 &  0.895 $\pm$ 0.160 &  0.854 $\pm$ 0.040 \\
\textbf{D9}      &     0.221 $\pm$ 0.054 &  0.353 $\pm$ 0.069 &          0.421 $\pm$ 0.093 &   0.776 $\pm$ 0.150 &  \textbf{0.181 $\pm$ 0.014} &  0.299 $\pm$ 0.137 &  0.237 $\pm$ 0.055 \\
\textbf{D10}     &     1.477 $\pm$ 0.279 &  0.633 $\pm$ 0.101 &          0.005 $\pm$ 0.002 &   0.524 $\pm$ 0.339 &  0.004 $\pm$ 0.001 &  \textbf{0.002 $\pm$ 0.001} &  0.010 $\pm$ 0.010 \\
\textbf{D11}     &     0.609 $\pm$ 0.238 &  0.511 $\pm$ 0.096 &          1.459 $\pm$ 0.704 &   0.607 $\pm$ 0.236 &  0.467 $\pm$ 0.146 & \textbf{ 0.423 $\pm$ 0.134} &  0.483 $\pm$ 0.170 \\
\textbf{D12}     &     1.945 $\pm$ 0.238 &  1.074 $\pm$ 0.176 &          1.746 $\pm$ 0.968 &   1.126 $\pm$ 0.366 &  0.482 $\pm$ 0.193 &  0.446 $\pm$ 0.149 &  \textbf{0.376 $\pm$ 0.094} \\
\textbf{D13}     &     5.3e1 $\pm$ 3.0e1 &  1.025 $\pm$ 0.118 &          1.295 $\pm$ 0.510 &   8.230 $\pm$ 7.666 &  1.010 $\pm$ 0.637 &  1.504 $\pm$ 1.544 &  \textbf{0.576 $\pm$ 0.313} \\
\textbf{D14}     &     1.0e1 $\pm$ 8.189 &  0.628 $\pm$ 0.168 &          1.734 $\pm$ 1.448 &   5.702 $\pm$ 3.840 &  1.961 $\pm$ 3.015 &  0.588 $\pm$ 0.438 &  \textbf{0.529 $\pm$ 0.394} \\
\textbf{D15}     &     0.278 $\pm$ 0.027 &  0.279 $\pm$ 0.042 &          0.352 $\pm$ 0.043 &   0.314 $\pm$ 0.064 &  0.257 $\pm$ 0.015 &  0.264 $\pm$ 0.026 &  \textbf{0.254 $\pm$ 0.022} \\
\textbf{D16}     &     1.576 $\pm$ 0.407 &  0.728 $\pm$ 0.111 &          4.732 $\pm$ 1.608 &   1.250 $\pm$ 0.201 &  0.716 $\pm$ 0.065 &  0.690 $\pm$ 0.033 &  \textbf{0.652 $\pm$ 0.092} \\
\textbf{D17}     &     2.5e2 $\pm$ 2.8e1 &  0.513 $\pm$ 0.015 &          2.6e3 $\pm$ 3.7e3 &   2.5e1 $\pm$ 6.141 &  0.157 $\pm$ 0.079 &  \textbf{0.052 $\pm$ 0.014} &  0.097 $\pm$ 0.010 \\
\midrule
\textbf{Average} &     2.2e1 $\pm$ 6.5e1 &  0.655 $\pm$ 0.296 &          7.8e2 $\pm$ 2.5e3 &   2.917 $\pm$ 5.911 &  0.443 $\pm$ 0.473 &  0.414 $\pm$ 0.357 &  \textbf{0.324 $\pm$ 0.227} \\
\bottomrule
\end{tabular}
    \label{tab:score}
    }}
    \end{center}
    \end{table} 
    
    % \newpage
\bibliographystyle{splncs04}
\bibliography{Refs/references,Refs/references1}
\end{document}